# REVIEW OF MACHINE LEARNING ALGORITHMS IN DIFFERENTIAL EXPRESSION ANALYSIS

**Irina Kuznetsova**

Graz University of Technology
Graz, AUSTRIA
and
Inst. of Interactive Systems and Data Science, Harry Perkins Inst. of Medical Research
University of Western Australia
Perth, AUSTRALIA

i.kuznetsova@hci-kdd.org

http://hci-kdd.org

**Yuliya V Karpievitch**

Plant Energy Biology Centre of Excellence
and
Harry Perkins Inst. of Medical Research
University of Western Australia
Perth, AUSTRALIA

yuliya.karpievitch@uwa.edu.au

**Aleksandra Filipovska**

Harry Perkins Inst. of Medical Research
University of Western Australia
Perth, AUSTRALIA

aleksandra.filipovska@uwa.edu.au

https://www.perkins.org.au/our-people/laboratory-heads/filipovska-profile/

**Artur Lugmayr**

Visualisation and Interactive Media Lab. (VisLab)
Curtin University
Perth, AUSTRALIA

lartur@acm.org

http://www.artur-lugmayr.com

**Andreas Holzinger**

Inst. of Interactive Systems and Data Science
Graz University of Technology
Graz, AUSTRIA

a.holzinger@hci-kdd.org

http://hci-kdd.org

## ABSTRACT

In biological research machine learning algorithms are part of nearly every analytical process. They are used to identify new insights into biological phenomena, interpret data, provide molecular diagnosis for diseases and develop personalized medicine that will enable future treatments of diseases. In this paper we (1) illustrate the importance of machine learning in the analysis of large scale sequencing data, (2) present an illustrative standardized workflow of the analysis process, (3) perform a *Differential Expression (DE)* analysis of a publicly available *RNA sequencing (RNA-Seq)* data set to demonstrate the capabilities of various algorithms at each step of the workflow, and (4) show a machine learning solution in improving the computing time, storage requirements, and minimize utilization of computer memory in analyses of RNA-Seq datasets. The source code of the analysis pipeline and associated scripts are presented in the paper appendix to allow replication of experiments.

## 1. INTRODUCTION

Every living cell's genome is encoded in *DNA (Deoxyribonucleic Acid)* – a long sequence of nucleic acids, also called nucleotides, encoded in four letters: *adenine*, *thymine*, *cytosine*, and, *guanine*, abbreviated as *A*, *T*, *C* and *G* respectively. DNA provides recipes for making all active molecules in the cell, such as *RNA (Ribonucleic Acid)* and proteins [1, 2]. In RNA *T* is replaced by *uracil (U)* and proteins are made of twenty amino acids.

Information stored in the DNA is transcribed into RNA some of which are further translated into proteins, which are the main workhorses of the cell. These three steps are also referred to as genome, transcriptome and proteome, which make up multidimensional data.

Currently genome and transcriptome are analysed via *Next Generation Sequencing (NGS)*. Sequencing is the process of transforming molecular information into a digital format. NGS allows sequencing the entire genome, specific regions of the genome, as well as epigenetic modifications of the genome (e.g. Methyl-C-seq, ChIP-seq) [3-6].

Sequencing RNA provides a snapshot of cellular gene expression, which is compared among various treatment groups or disease states to identify differentially expressed genes. *RNA sequencing (RNA-Seq)* reveals the order of the four nucleotides in short segments [7, 8]. These short segments are called *reads* and are stitched together algorithmically into a large genome sequence computationally [4, 6, 9]. There are two ways to sequence DNA/RNA fragments using single-end or paired-end sequencing. In single end sequencing DNA/RNA fragment is sequenced from one end only, whereas in pair-end sequencing fragment is sequenced from two opposing ends producing two reads per fragment.

NGS generates vast amounts of data, usually hundreds of gigabytes [5], commonly referred to as Big Data, and as researchers we aim to transform into Cognitive Big Data [10, 11] or other practical use cases as e.g [58] or [59]. Therefore automatic machine learning plays a crucial role in handling, interpreting, learning and visualising the big NGS datasets to produce easily understandable knowledge base.

Here we illustrate the importance of machine learning algorithms in analyzing big data and provide a specific example of analysis pipeline (also referred to as workflow) of NGS data. Our analysis pipeline consist of a standardised modular workflow where some modules are taken from the pipeline proposed by Partea et al. [12].

Within the scope of this paper, we show the utility of machine learning algorithms in identifying genes that are different among various conditions. We use two publicly accessible mouse RNA-Seq data sets available from the NCBI GEO database accession NCBI GEO database under the accession number GSE56933 [13] and under the accession number GSE60450 [14].

### a. Research Goals

Here we provide a detailed review of the algorithms most widely used in RNA-Seq DE analysis. We showcase the necessity and requirements for each step in the analysis pipeline and deliver the minimal required knowledge about RNA-Seq DE analysis.

We show:

- practical impossibility of analysing NGS data without the use of computer algorithms
- standardized workflow of DE analysis
- utilization of publicly available data for analysis and new algorithm development

To achieve our goal, we first describe several potential pipelines. Next, we evaluate and selected software and last, we perform an illustrative performance analysis which can be used as a learning example.

## 2. TOOLS FOR TRANSCRIPTOMIC DATA ANALYSIS

Transcriptome studies using RNA-Seq reveal quantitative information about transcripts, which are fragments of RNA. Transcripts, in turn, show variation in patterns of gene expression at specific developmental time points, in various treatment conditions [9, 15-17]. As such, the most common goal in RNA-Seq analysis is finding differentially expressed genes across different conditions.

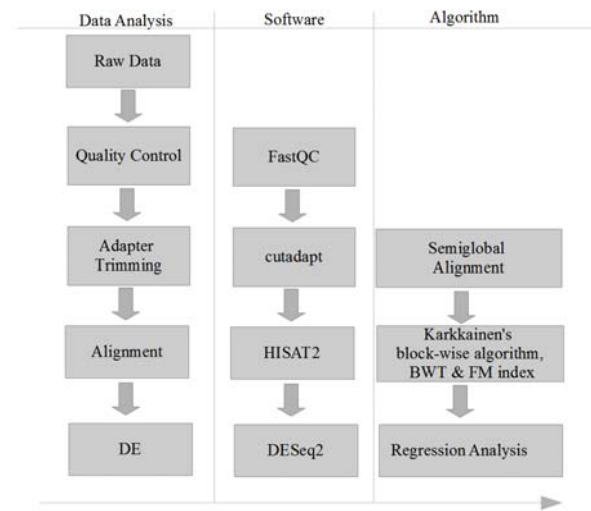

**Figure 1. Data analysis workflow, software, algorithms.**

A standard RNA-Seq workflow can be divided into the following steps:

1. Design of experiment: determine the biological question to study and estimate the sample size necessary to identify new knowledge with statistical significance
2. Perform the biological experiment using cells or animal models under different treatment conditions or genetic alterations
3. Obtain sequencing data [7]

4. Preprocess data: perform quality control, adapter trimming, and alignment
5. Analyze data: identify coverage at gene- or transcript- expression level, normalize, find which genes behave differently across conditions, visualize results, and validate results if needed

Our differentially expression analyses example is outlined in Figure 1. It includes quality control, adapter trimming, alignment, and differential expression. The tools we suggest to use are FastQC [18] (quality control), cutadapt [19] (adapter trimming), HISAT2 (alignment) [20], DESeq2 [21] R package (DE). There are alternative methods to perform most of the steps.

To understand the performance of the applied tools, we conducted a performance analysis such as algorithm execution time, memory (RAM) and CPU usage.

Table 1. Representation of tools that can be used for RNA-Seq data analysis.

| | | Task | Tools | Algorithm | Ref |
|---|---|---|---|---|---|
| **Preprocessing** | Quality control | Adapter trimming; poor quality bases elimination | Cutadapt | Semiglobal alignments | [19] |
| | | | FASTX-Toolkit | - | [22] |
| | | | Trimmomatic | Seed and extend followed by palindrome mode approach | [23] |
| | | | BBDuk | - | [24] |
| | Alignment | Alignment of reads to the reference genome | BWA | Burrows-Wheeler transform | [25] |
| | | | Bowtie 2 | Burrows-Wheeler transform | [26] |
| | | | HISAT2 | Karkkainen's blockwise algorithm | [20] |
| **Analysis** | Differential Expression | Identify differentially expressed genes | Deseq2 (R) | Negative binomial generalized linear models | [21] |
| | | | edgeR (R) | Negative binomial generalized linear models | [27] |
| | | | Ballgown (R) | Standard linear model-based comparison statistical test | [28] |

## 3. PIPELINE SETUP

### a. Data Description

We downloaded the GSE60450 [14] dataset from the *Gene Expression Omnibus (GEO)*. These data are of gene expression changes in luminal and basal mammary glands of non-pregnant, pregnant and lactating mice was run on the Illumina HiSeq 2000 platform. Sequenced total RNA from the samples generated single-end reads of 100 *base pairs (bp)* in length.

The second GSE56933 [13], RNA-Seq dataset was extracted from the heart and liver tissues of 10-weeks old male mice and ran on the Illumina Genome Analyzer IIx. The resulting reads are single-end of 75 bp in length and contain adapter sequences. These data are selected to exemplify the necessity of adapter trimming algorithms that remove adapter sequences before the data is aligned to a reference genome. Since adapter sequences are not present in organism's genome reads that contain sequences will fail to correctly align to the genome.

In both datasets raw sequence information was saved in FASTQ file format. FASTQ file format is text-based with four lines of the file describing one read/sequence at a time. The first line is a sequence identifier, the second, is the sequence itself, the third line is no longer used for sequence identification and contains a +, and forth line contains sequence quality information [29].

### b. Quality Control

Quality control is performed on the raw data to detect low quality bases, duplicates, PCR primers, or adapters in the reads. FastQC [18], an open-source software is widely used to investigate: (1) per base sequence quality, (2) per sequence quality scores, (3) per base sequence content, (4) per base GC content, (5) per base N content, (6) sequence length distribution, (7) duplication level, (8) overrepresented sequences, (9) adapter content, (10) kmer content, and (11) per tile sequence quality [18]. The quality summary provides information on possible artifacts in the raw data that can affect the next steps of the RNA-Seq analysis. The adapter content section of the FastQC report gives information about the adapter sequence observed in the data. Additionally, the overrepresented sequences section of the FastQC report sometimes, but not always, gives more specific information of the possible adapter source. For example, Figure 2A shows a list of overrepresented sequences found in GSE56933 dataset, and indicates presence of the TruSeq adapter. The pattern of the adapter can be found in the Illumina adapter catalogue and should always be provided by the authors [30].

| A. Overrepresented sequences before adapter trimming | | | |
|---|---|---|---|
| Sequence | Count | Percentage | Possible Source |
| GATCGGAAGAGCACACGTCTGAACTCCAGTCACGAGTGGATATCTCGTATGCCGTCTTCTGCTTGAAAAAAAAAA | 673591 | 2.695638929 | TruSeq Adapter, Index 7 (97% over 36bp) |
| AGATCGGAAGAGCACACGTCTGAACTCCAGTCACGAGTGGATATCTCGTATGCCGTCTTCTGCTTGAAAAAAAAA | 394423 | 1.578438538 | TruSeq Adapter, Index 7 (97% over 36bp) |
| GATCGGAAGAGCACACGTCTGAACTCCAGTCACGAGTGGATATCTCGTATGCCGTCTTCTGCTTGAAAAAAAACA | 96534 | 0.386318713 | TruSeq Adapter, Index 7 (97% over 36bp) |
| GATCGGAAGAGCACACGTCTGAACTCCAGTCACGAGTGGATATCTCGTATGCCGTCTTCTGCTTGAAAAAAAAAC | 78364 | 0.313604322 | TruSeq Adapter, Index 7 (97% over 36bp) |
| GATCGGAAGAGCACACGTCTGAACTCCAGTCACGAGTGGATATCTCGTATGCCGTCTTCTGCTTGAAAAAAACAA | 60850 | 0.243515173 | TruSeq Adapter, Index 7 (97% over 36bp) |
| AGATCGGAAGAGCACACGTCTGAACTCCAGTCACGAGTGGATATCTCGTATGCCGTCTTCTGCTTGAAAAAAACA | 42623 | 0.170572674 | TruSeq Adapter, Index 7 (97% over 36bp) |
| AGATCGGAAGAGCACACGTCTGAACTCCAGTCACGAGTGGATATCTCGTATGCCGTCTTCTGCTTGAAAAAAAAC | 37795 | 0.151251536 | TruSeq Adapter, Index 7 (97% over 36bp) |
| GATCGGAAGAGCACACGTCTGAACTCCAGTCACGAGTGGATATCTCGTATGCCGTCTTCTGCTTGAAAAAAAATA | 31804 | 0.127276197 | TruSeq Adapter, Index 7 (97% over 36bp) |
| GATCGGAAGAGCACACGTCTGAACTCCAGTCACGAGTGGATATCTCGTATGCCGTCTTCTGCTTGAAAAAAAAGA | 27342 | 0.109419751 | TruSeq Adapter, Index 7 (97% over 36bp) |
| GATCGGAAGAGCACACGTCTGAACTCCAGTCACGAGTGGATATCTCGTATGCCGTCTTCTGCTTGAAAAAAAAAG | 26938 | 0.107802987 | TruSeq Adapter, Index 7 (97% over 36bp) |
| GATCGGAAGAGCACACGTCTGAACTCCAGTCACGAGTGGATATCTCGTATGCCGTCTTCTGCTTGAAAAAAAAAT | 26640 | 0.106610422 | TruSeq Adapter, Index 7 (97% over 36bp) |
| | | | |
| B. Overrepresented sequences after adapter trimming | | | |
| Sequence | Count | Percentage | Possible Source |
| | 2017092 | 8.072185821 | No Hit |
| A | 934711 | 3.74061316 | No Hit |

Figure 2. FastQC HTML report overview of overrepresented sequences before (A) and after (B) trimming [18].

c. **Adapter Trimming**

Once the presence of adapters is identified, the next step is to remove/trim those adapters prior to read alignment to a reference genome.

To perform adapter trimming we used one of the commonly used trimming tools, cutadapt [19]. Cutadapt can remove adapters an error-tolerant way, which means that it will trim adapters even if errors were introduced during the sequencing. The adapter sequence can be fully or partially present in the read.

Cutadapt uses unit costs function to consider mismatches (a single nucleotide base substitution), insertions (one or more base pair are added to the sequence), or deletions (one or more base pairs are lost in the sequence) as a single error score. The best mapped sequences are those that have an overlap score maximized but are below the selected threshold error rate. If after passing this condition there are multiple options of mapped sequences, the mapping with the smallest error rate is selected. However, if by passing this condition there are still more than one mapped sequences, then the mapping of the adapter sequence to the most left position on the read is selected as the best match [19]. To validate that all adapters have been trimmed we run FastQC software on adapter trimmed data. Figure 2B shows a list of overrepresented sequences after trimming with cutadapt. Table 1 contains a list of adapter trimming software.

d. **Alignment**

The next step in the analysis pipeline is read alignment. For organisms for which genome sequence (reference genome) is available the reads are aligned to that organism's reference genome. For organisms without genome reference genome *de novo* assembly is required (de novo assembly is beyond the scope of this manuscript).

Although there are many programs that can perform sequence alignment not all of them are appropriate for NGS data, such as BLAST [31] and BLAT [32], as these tools were developed for low volume datasets.

Table 1 contains a list of aligners suitable for NGS sequencing data. The main goal of the alignment process is to correctly align and assemble a large number of short reads to a reference genome, which is a time consuming process, requires temporary disk storage, and intensive CPU usage. Majority of currently available alignment tools use algorithms that can be categorized into two groups: (1) based on hash tables and (2) based on suffix tree [33].

The hash table algorithm stores reads of the data in an array that can be retrieved with an index, where similar values are stored at the same location under the same index. On the other hand, the suffix array keeps the suffixes of the data values in a tree-like manner, where identical copies of reads' suffixes are stored at the same path. Both approaches attempt to minimize execution time.

When selecting an algorithm to process the data one has to consider the following factors: (1) reference genome size, for example, human genome consists of approximately of three billion bp [34], which comes at a cost of time and space for its alignment as compared to a shorter genome, (2) the length of the individual read can range from 25 to 500 bp and the length affects the accuracy and speed of the alignment. The longer reads will take longer to align.

In our example we utilize *Hierarchical Indexing for Spliced Aligner (HISAT2)*[16] that uses the block-wise algorithm of Karkkainen in combination with the *Burrows-Wheeler Transformation (BWT)* and *Ferragina and Manzinin (FM)* index, data transformation and compression algorithms respectively [35-37]. HISAT2 [16, 20] can be used to align genes that have annotated splice sites, unlike Bowtie 2 [26], which is splice-unaware aligner. Figure 3 depicts the need for splice-aware alignment in higher level organisms.

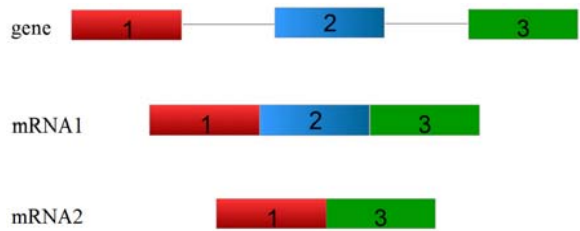

**Figure 3. Alternative splicing. One gene with three introns is transcribed into two different mRNAs, one containing all three introns (mRNA1), and the second (mRNA2) with the second intron spliced out.**

The BWT is a compression algorithm and *suffix array (SA)* is lexicographically sorted array that when combined create space-efficient index or the FM-Index, which uses a prefix as a search pattern [36]. Storage space requirements of SA can be reduced to small blocks that is the approach of Karkkainen's algorithm [37]. A detailed explanation of the BWT and FM-indexes is described in Langmead's tutorial *("Introduction to the Burrows-Wheeler Transform and FM Index" Langmead 2013)* and the block-wise algorithm is thoroughly explained in [37].

The output of the alignment software is a *SAM* format file which is a tab delimited file that contains the alignment information, for example, a read sequence that mapped to a genome, quality score of the alignment, genome mapping position (coordinate) [38]. The SAM files are large human readable text files, which require a lot of storage space, therefore they are commonly converted to a binary BAM format [38]. BAM files are accompanied by index files with extension *bai* to speed up access time.

All aligners will produce a short summary of the alignment such as a number of raw reads and the number of aligned reads. If the number of aligned reads is satisfactory, which is generally anywhere above 70% of raw reads, next step is to visualise the aligned reads. Genome browser such as *Integrative Genomics Viewer (IGV)* [39] or UCSC genome browser [40] are some of the best ways of confirming correct alignment.

IGV is a freely available Java based visualization platform that is run locally on a computer, and enables to explore fragments that are mapped to the reference genome. IGV interface enables one to perform wide range of tasks, such as zooming in to the region of chromosome/gene of interest with good resolution, coloring, or sorting [39]. Due to the need of installation on a local PC performance of IGV is faster than web-based genome browsers (such as the UCSC genome browser). IGV is a quick way to visualize the data, whereas researchers use UCSC browser to produce publication qualities images. UCSC browser allows for easy data sharing among multiple collaborators.

Figure 4 illustrates read coverage (raw counts) for the *EGF* gene that provides a general overview of the depth of sequencing coverage. Visualization of the entire genome coverage provides limited information due to data being highly condensed; zooming in at the level of a region or gene of interest will provide sufficient detail about the coverage.

e. **Differential Expression**

DE analysis investigates differences in RNA levels among various samples as readout of gene expression changes. In general, DE analysis takes aligned raw counts, subjects them to normalization to improve comparability across samples prior to estimating statistical significance of the gene expression change.

Several R packages exist that analyze RNA-Seq data for detecting differentially expressed genes across various conditions. The most popular are edgeR [26], DESeq2 [21] and Ballgown [16, 27]. While DESeq2

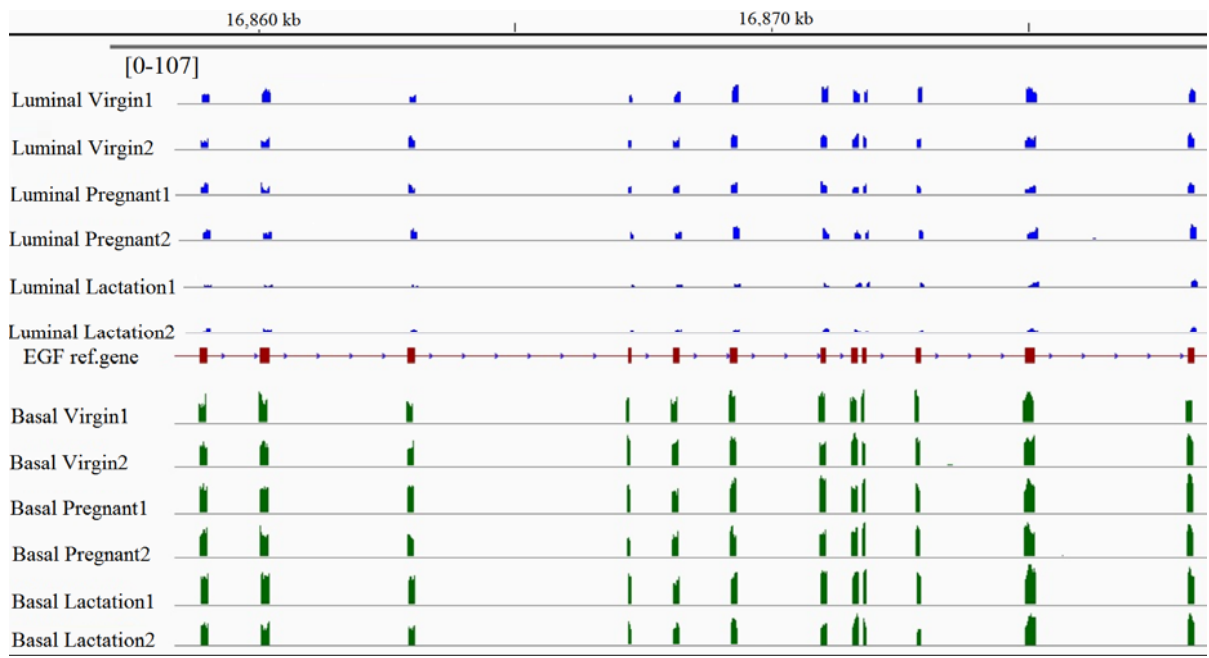

**Figure 4. Coverage representation of aligned reads. IGV snapshot of EGF gene.**

[21] and edgeR [26] are used for DE analysis based on gene annotations and are similar in performance, Ballgown [16, 27] is capable of analyzing both gene and transcript annotations. Both edgeR and DESeq2 packages take raw counts as input data [40]. Ballgown requires transcript assembly with StringTie to be performed, which stores results in *ctab* format as gene-, transcript-, exon- and junction-level expression measurements. All methods perform count normalization to the total number of reads (library size) to produce abundance estimates [16, 21, 26, 27].

*Statistical analysis*

Both DESeq2 and edgeR require at least three samples per treatment group, whereas it is suggested to use Ballgown with four or more replicates due to the linearity in the model-based analysis [16]. We chose to perform DE analysis of annotated genes without the need to identify novel isoforms. For this reason and the availability of three biological samples per variable the most suitable package for DE in our example was DESeq2. The raw counts matrix was extracted with the provided python script (http://ccb.jhu.edu/software/stringtie/dl/prepDE.py) and supplied and analyzed with DESeq2 [21] (Appendix). The choice of DESeq2 enabled us to do the analysis quickly, exemplifying that the choice of the program and algorithms can cater to the needs of the biological question and sample availability.

*Visualization*

Visualization is an important step in data analysis that allows for ease of data interpretation. There are many possible ways of presenting the results of expression data such as MA-plot, volcano plot, counts plot, or heatmap among others. *Clustering Image Map (CIM)* visualized as a heatmap shows differences and similarities across various conditions. Figure 4 shows the most differentially expressed genes of EGF receptor family.

Heatmap is a two-dimensional display of numbers as colors. Heatmap has rows that indicate genes, and columns that show different conditions or samples. Figure illustrates a heatmap of 11 pre-selected significantly expressed genes of EGF receptor family under three conditions (virgin vs pregnant vs lactation) for two cell types (luminal vs basal). The color key scheme (top left) shows the association of the numeric values to the color scheme. Here lowly expressed genes are shown in blue, and highly expressed genes are shown in red.

Most heatmap functions can perform hierarchical clustering of genes and samples, which are shown as a dendrogram on the left and top of the plot in Figure 5.

Here rows/genes that have similar gene expression patterns are grouped together, and in columns luminal and basal cells represent two separate clusters. Moreover, the pregnant and virgin conditions show similar pattern in gene expression as compared to lactation condition for luminal cell type. The fact that two replicates for each cell type/condition cluster together confirms good quality of our example data. Heatmaps allow one to employ different algorithms to clustering the data, which will effect the dendogram.

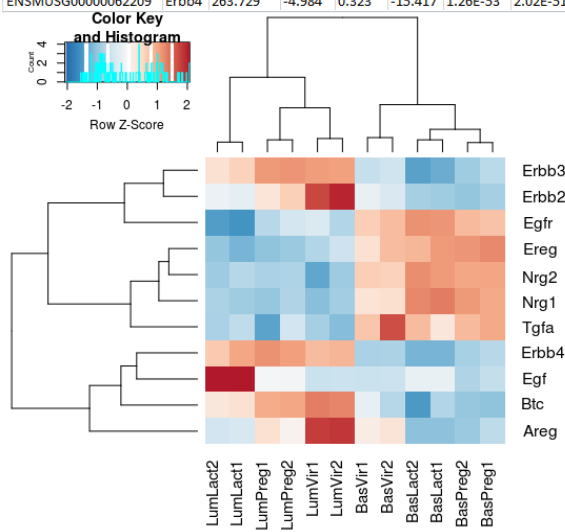

**Figure 5.** DESeq2 resulting table for EGF receptor family (top panel) and heatmap of the most differentially expressed genes of EGF receptor family (bottom panel).
*Note: only three digits are shown after decimal point.

## 4. PRESENTATION AND PERFORMANCE ANALYSIS

Within this section, we describe a method for conducting a performance analysis for the standard DE analysis for RNA-Seq data. Table 2 illustrates the most popular tools in the domain, including the software versions that we have been applying as part of our DE analysis.

To evaluate the performance of the applied software tools, we utilized the performance counter tool (*perf*) [41]. Perf is a Linux based command line utility that provides performance information of the operating system, applications and hardware. *Perf stat* function is an excellent choice for software that can provide performance analysis.

**Table 2. Software Tools Applied for a DE Analysis**

| Software Tool | Version | Ref |
|---|---|---|
| FastQC | v0.11.5 | [18] |
| Cutadapt | v1.10 | [19] |
| HISAT2 | v2.0.4 | [20] |
| DESeq2 (R) | v3.3.2 | [21] |
| Perf (Linux) | v4.9.rc8.g810ac7b7 | [41] |

Table 3 illustrates performance parameters for FastQC, cutadapt, HISAT2 as *Instructions per Cycle (IPC)*, where results are represented as software and hardware related events. The task-clock, context-switches, cpu-migrations, page-faults are software related events, and the cycles, instructions are related to the hardware events. The task-clock shows the amount of time spend on the task. However, if there is parallel computing involved (for example, threads option in HISAT2) this number has to be devised on the CPUs involved. Context switch explains how many times the software switched of the CPU from one process/thread to another. CPU migration describes equality in a work load distribution across all cores. Finally, the page-faults occur when a program's virtual content have to be copied to the physical memory.

Our initial tests show, that HISAT2 (without threads option) is one of the most time-consuming steps (~25 min) in the analysis process, with the highest CPU usage (1.047 CPUs utilized). This can be explained due to the nature of the alignment process, where enormous amount of reads are required to be mapped the reference genome. Although HISAT2 utilizes Karkkainen's algorithm, which is time and space efficient, volume of the data that is required to be mapped to the reference genome is enormous and takes time to execute. One of the solutions to this problem is to apply threads (-p) option that enables to process the data in a parallel. We run HISTA2 with 18 threads which reduced the run time to 5 minutes. The number of cycles, which are an indicator for the number of instructions performed by software to produce final result is the highest for HISAT2 with thread parameter (~2.791 GHz), as expected, due to the fact that multiple parallel processes were executed.

Table 3. Perf Software Performance Analysis on the Example of FastQC, cutadapt, and HISAT2 utilizing the (SRR1552444) Data Set. Data is presented in Instructions per Cycle (IPC)

| Performance Parameter | FastQC | Cutadapt | HISAT2 | HISAT2 with thread –p 18 | Measured |
|---|---|---|---|---|---|
| **Software Events** | | | | | |
| task-clock | 3.8 (min) / 1.021 CPUs utilized | 5.6 (min) / 0.995 CPUs utilized | 25.9 (min) / 1.047 CPUs utilized | 70.28 (min) / 14.534 CPUs utilized | |
| context-switches | 0.094 | 0.002 | 0.752 | 4e-6 | K/sec |
| cpu-migrations | 0.006 | 0.000 | 0.001 | 0.011 | K/sec |
| page-faults | 0.302 | 0.019 | 3.6e-5 | 1.3e-5 | K/sec |
| **Hardware Events** | | | | | |
| cycles | 2.925 | 2.976 | 2.959 | 2.791 | GHz |
| Total number of cycles | 6.70122e+11 | 1.00997e+12 | 4.61275e+12 | 1.17684e+13 | |
| instructions | 1.14 | 1.27 | 0.91 | 0.48 | insn per cycle |
| **Total Run Time** | | | | | |
| | 3.7 | 5.7 | 24.8 | 4.8 | minutes |

## 6. CONCLUSIONS AND DISCUSSION

Within the scope of this paper we contributed with:

- evaluation of machine learning algorithms utilized for differential expression analysis of RNA-Seq data
- an example pipeline, which can be used to perform DE analysis (Figure 1)
- source code of a script-based pipeline

### a. Advantages of Applying Machine Learning

The application of machine learning helps to analyze large volumes of data without loading it into the computer memory. The initial size of the SRR1552444.sra sample (GEO GSE60450 [14]) was 9.7 GB was narrowed to 1.1 MB of meaningful data, in other words we identified 7170 of significantly expressed genes with only 15 minutes software run time.

### b. Source Code of the Script Based Workflow

Appendix A contains the source code and a practical guideline for people interested in conducting a similar kind of analysis. We provided and evaluated a script workflow of an analysis process to allow others repeating the experiment.

### c. Generic Reference Pipeline and Workflow

To conduct the analysis, we applied publicly available datasets with the accession number GSE56933 [13]; and accession number GSE60450 [14] to describe the performance of software used in DE analysis of limited (up to three samples) number of samples.

The analysis process contains four steps:

- quality control
- adapter trimming
- alignment
- differential expression analysis

This is a standard workflow for DE analysis and will only vary due to selection of various algorithms for each step. For example, adapter trimming step may be omitted if the data has been already preprocessed in the past and was downloaded from GEO.

Here we show how to obtain data from publically accessible repositories can be used for analyses. One can also use such data to develop novel algorithms that can be benchmarked against existing ones without the need to produce more sequencing data. Algorithms described here provide efficient processing of big data but further improvements in speed, disk and RAM utilization are necessary to deal with larger and larger datasets.

### d. Final Remarks

Big data coming from various omics platforms will only increase in size in the near future thus increasing the requirements for high performance analysis to gain understanding of the data. To process, visualize, and understand such omics big data we must apply machine learning algorithms. Interestingly, and depending on the complexity of the task, multiple algorithms are utilized and parametrized to improve performance. Within this paper, we gave an overview several widely used analysis algorithms for NGS data, and indicated steps that assist in the analysis process.

Future work and development will require improved machine learning algorithms such as deep learning on distributed systems [42, 43]. In addition porting existing statistical methods from other omics platforms [44-48] as well as developing new ones [16, 28, 49] specific to NGS is necessary and will provide greater statistical certainty in the results.

In biomedical sciences, however, automatic machine learning approaches may suffer from insufficient number of training samples as a result of limited number of biological data sets, and in this instance *interactive Machine Learning (iML)* may be particularly useful [50, 51]. A grand challenge is to provide integrative machine learning approaches, i.e. the optimization of workflows and processes that are in-line with the main workflow of biomedical researchers, thereby increasing their capacity whilst reducing costs and improving efficiency [52, 53]. In this context usability gets a new meaning and an increasing importance, as experimental scientists often have limited skills in machine learning generally or in algorithms specifically – raising the need for multidisciplinary training the next generation of researchers in biology, bioinformatics and statistics [54].

Finally, bioinformatics software needs to be user-friendly and be accompanied by a comprehensive user manual. User-friendly, well documented software we provide to biologists will ease the discovery process in biological sciences to improve our understanding of diseases [55, 56] and transform our medical system into truly personalized medicine [57].

APPENDIX: SOURCE CODE AND SCRIPTS TO CONDUCT A SIMILAR KIND OF ANALYSIS

Herein we describe step by step procedure used to achieve final results of the DE analysis to make the analysis process available to others. The Unix shell is used to run majority of commands of described protocol, Python, and R. One sample is taken to show how to use these tools. We tried to present each step in a very precise way by avoiding one line complex commands

ANALYSIS PROCESS GUIDELINES FOR SOFTWARE INSTALLATION

**NCBI SRA Toolkit**:
https://trace.ncbi.nlm.nih.gov/Traces/sra/sra.cgi?cmd=show&f=software&m=software&s=software
**FastQC**:         http://www.bioinformatics.babraham.ac.uk/projects/fastqc/
**Catadapt**:       http://cutadapt.readthedocs.io/en/stable/installation.html
**HISAT2**:         https://ccb.jhu.edu/software/hisat2/manual.shtml
**Samtools**:       http://www.htslib.org/download/
**StringTie**:      https://ccb.jhu.edu/software/stringtie/#install
**Python** (prepDE.py): http://ccb.jhu.edu/software/stringtie/index.shtml?t=manual
**R** (Deseq2):     source("https://bioconductor.org/biocLite.R")
                    biocLite("DESeq2")

ANALYSIS PROCEDURE

STEP 1: RAW DATA

Note: the first two steps are performed for GSE56933 dataset based on one sample (SRR1257444).

Download *sra* files. Unix shell is used to download the samples data from NCBI Geo database. The ftp directory of each sample should be supplied to *wget*. The samples will be saved to the directory from which the command is run. Run following command to download sample:

```
$ wget ftp://ftp-trace.ncbi.nlm.nih.gov/sra/sra-
instant/reads/ByExp/sra/SRX/SRX681/SRX681985/SRR1552444/ SRR1552444.sra
```

The next step is to convert *sra* to *fastq* file format. This can be done with *sra-toolkit* (see software installation section):

```
$ path_to_sratoolkit/fastq-dump –gzip –split-3 SRR1257444.fastq
```

STEP 2: QUALITY CONTROL

Quality control with *FastQC*:

```
$ fastqc -o SRR1257444.fastq.gz
```
Unzip file:
```
$ gunzip SRR1257444.fastq.gz
```

STEP 3: ADAPTER TRIMMING

Adapter trimming with *cutadapt*. Use only the prefix of the adapter sequence (TruSeq Index):

```
$ cutadapt -a  GATCGGAAGAGCACACGTCTGAACTCCAGTCAC -o SRR1257444.fastq
adapt_tr_SRR1257444.fastq &> stat_SRR1257444.log
```

## STEP 4: ALIGNMENT

Note: following steps are performed for data set from GSE60450, sample (SRR1552444). Map reads to the mouse reference genome:

```
$ hisat2 -p 18 --dta -x index/mouse_genome -U SRR1552444.fastq -S
aligned_SRR1552444.sam &> stat_SRR1552444.log
```

Sort and convert the *SAM* to *BAM*:

```
$ samtools sort -@ 18 -o SRR1552444.bam SRR1552444.sam
```

Generate *BAM* index file:

```
$ samtools index SRR1552444.bam
```

Assemble transcripts.

```
$ stringtie -p 18 -e -B -G musmusc.gtf -o outfolder/SRR1552444/SRR1552444.gtf -l
SRR1552444 SRR1552444.bam
```

Note: Important the *outfolder* holds all sample output from *stringtie*. For example, for the next sample the *stringtie* command line looks like:

```
$ stringtie -p 18 -e -B -G musmusc.gtf -o outfolder/SRR1552445/SRR1552445.gtf -l
SRR1552445 SRR1552445.bam
```

### PYTHON

Follow an alternative differential expression workflow. Download *prepDE.py* from http://ccb.jhu.edu/software/stringtie/index.shtml?t=manual. Note that *prepDE.py* script aims to extract raw count matrixes for gene-level measurements and transcript-level measurements. The goal of this paper is to show differentially expression analysis at gene-level, thus we use only *gene_level_raw_counts.csv* output file. The script should be saved to the same directory as the folder that holds *stringtie* output (same path as *outfolder*)

## STEP 5: DIFFERENTIAL EXPRESSION ANALYSIS

### R

Load *DESeq2* package:

```
$ R
R version 3.3.2 (2016-10-31)
source("https://bioconductor.org/biocLite.R")
biocLite("DESeq2")
library(DESeq2)
```

Set directory to a file. Load gene-level count matrix:

```
setwd("path to gene_level_raw_counts.csv")
file_count = read.csv("gene_level_raw_counts.csv",row.names=1)
countData = as.matrix(file_count)
```

Create phenotype data ( multi-compariosn):

```
ctype=factor(c(rep('Luminal',6),rep('Basal',6)))
condition=factor(c(rep('Virg',2), rep('Pregn',2), rep('Lact',2)))
coldata <- data.frame(row.names=colnames(countData),ctype,condition)
```

Construct DESeq object:

```
dds = DESeqDataSetFromMatrix(countData = countData, colData = coldata, design = ~condition + ctype)
```

Filter law abundance genes:
```
dds = dds[ rowSums(counts(dds))>1,]
```

Differential expression analysis based on the Negative Binomial distribution:

```
dds = DESeq(dds)
res = results(dds, contrast=c("ctype","Basal","Luminal"))
```

Get significantly expressed genes (cut-off < 0.01)

```
signif = res[res$padj<0.01 & !is.na(res$padj),]
write.csv(as.data.frame(signif),file="DESeq2_signif_two_group_gene_level_001.csv")
```